\documentclass{article}

\usepackage[preprint,nonatbib]{neurips_2022}

\usepackage[utf8]{inputenc} 
\usepackage[T1]{fontenc}    
\usepackage{hyperref}       
\usepackage{url}            
\usepackage{booktabs}       
\usepackage{amsfonts}       
\usepackage{nicefrac}       
\usepackage{microtype}      
\usepackage{xcolor}         

\usepackage{times}
\usepackage{epsfig}
\usepackage{graphicx}
\usepackage{amsmath}
\usepackage{amssymb}

\usepackage{subfigure}

\usepackage{algorithm}
\usepackage{algorithmic}

\usepackage{placeins}
\usepackage{multirow}

\title{Learning Convolutional Neural Networks in the Frequency Domain}

\author{%
	Hengyue Pan \\
	School of Computer\\
	National University of Defense Technology\\
	109 Deya Road, Changsha, China 410073 \\
	\texttt{hengyuepan@nudt.edu.cn} \\
	 \And
	 Yixin Chen \\
	 School of Computer\\
	 National University of Defense Technology\\
	 109 Deya Road, Changsha, China 410073 \\
	 \texttt{chenyixin@nudt.edu.cn} \\
	\And
	Xin Niu \\
	School of Computer\\
	National University of Defense Technology\\
	109 Deya Road, Changsha, China 410073 \\
	\texttt{niuxin@nudt.edu.cn} \\
	\And
	Wenbo Zhou \\
	School of Computer\\
	National University of Defense Technology\\
	109 Deya Road, Changsha, China 410073 \\
	\texttt{zhouwenbo2021@nudt.edu.cn} \\
	\And
	Dongsheng Li \\
	School of Computer\\
	National University of Defense Technology\\
	109 Deya Road, Changsha, China 410073 \\
	\texttt{dsli@nudt.edu.cn} \\
}

\begin{document}
	
	\maketitle
	
	\begin{abstract}
		Convolutional neural network (CNN) has achieved impressive success in computer vision during the past few decades. The image convolution operation helps CNNs to get good performance on image-related tasks. However, the image convolution has high computation complexity and hard to be implemented. This paper proposes the CEMNet, which can be trained in the frequency domain. The most important motivation of this research is that we can use the straightforward element-wise multiplication operation to replace the image convolution in the frequency domain based on the Cross-Correlation Theorem, which obviously reduces the computation complexity. We further introduce a Weight Fixation mechanism to alleviate the problem of over-fitting, and analyze the working behavior of Batch Normalization, Leaky ReLU, and Dropout in the frequency domain to design their counterparts for CEMNet. Also, to deal with complex inputs brought by Discrete Fourier Transform, we design a two-branches network structure for CEMNet. Experimental results imply that CEMNet achieves good performance on MNIST and CIFAR-10 databases. 
	\end{abstract}
	
	\section{Introduction}
	
	In the past few decades, CNN \cite{lecun1995convolutional} has played an essential role in the field of computer vision. Even though ViT \cite{dosovitskiy2021an} shows excellent performance on vision tasks, CNN is still an irreplaceable tool for researchers and real-world users. The core of CNN is the image convolution operation, which is very complicated and hard to be implemented and parallelized. Based on theorems of signal processing, image convolution can be replaced by the straightforward element-wise multiplication via converting input data into the frequency domain using Discrete Fourier Transform (DFT) (see Section~\ref{Math} for more details). In this way, we may simplify the forward and backward calculations of convolution layers and make them easier to be parallelized. Moreover, \cite{trainingf} implies that CNNs may firstly capture low-frequency components of features, then high-frequency components. In \cite{highfreqcomponent}, authors emphasized that high-frequency components play important roles in the generalization of CNNs, even though they are almost imperceptible to human beings. Those researches imply that directly learning CNNs in the frequency domain may allow us to learn features of low-frequency and high-frequency components explicitly, then help us find more information to explain how CNNs work. 
	
	There are many efforts to implement CNNs in the frequency domain. \cite{LeCunDFTDNN2014} is an early research that considered to do convolution operation in the frequency domain. During the forward pass, the algorithm firstly transfers input feature maps and convolution filters into the Fourier domain, then applies the element-wise multiplication between them instead of the image convolution. After that, the result feature maps should be moved back to the time domain. The method achieved good efficiency, but it could not learn convolution filters directly in the frequency domain. \cite{Vasilache2015FastCN}, \cite{FastCNN}, and \cite{FALCON} mainly focused on the inference stage. Those researches implied that transferring well-trained CNNs into the frequency domain may obviously speed-up during the inference stage. 
	
	In \cite{NIPS2015_536a76f9} and \cite{DFTPooling}, DFT was applied on pooling layers of CNNs and result in good image classification performances. \cite{WATANABE2021107851} analyzed working behaviors of convolution layers, Batch Normalization, and ReLU in the frequency domain, and proposed a novel ReLU layer called 2SReLU that can work in the frequency domain. A convolution-free neural network with 2SReLU layers was trained in the frequency domain using MNIST and AT\&T databases. \cite{FCNN} proposed a Fourier Convolution Neural Network (FCNN), which includes Fourier Convolution Layers and Fourier Pooling Layers. FCNN can be trained entirely in the frequency domain using MNIST and CIFAR-10 databases. Different from the works above, \cite{RethinkingFUN} used Discrete Cosine Transform (DCT) to covert features to the frequency domain and result in a model family with good performance.   
	
	In this paper, we propose a novel computation model, namely CEMNet (Complex Element-wise Multiplication Network). CEMNet can be viewed as a counterpart of CNNs, which mainly works in the frequency domain. We provide theoretical analyses of convolution layers, Batch Normalization, Leaky ReLU and Dropout in the frequency domain, and propose corresponding layers to construct the CEMNet. Moreover, we design a two-branches structure to implement the CEMNet for complex inputs derived by DFT. The experimental results on MNIST and CIFAR-10 databases show that the proposed CEMNet has good performances on image classification tasks and less number of operations. The main contributions of this paper include: 
	
	(1) We propose the CEMNet that can be trained directly in the frequency domain, and using an element-wise multiplication to replace the image convolution operation of CNNs to reduce the computation complexity. Moreover, we introduce a Weight Fixation mechanism to deal with the over-fitting problem.
	
	(2) We implement Batch Normalization, Leaky ReLU, and Dropout in the frequency domain to improve the performance of CEMNet, and design a two-branches structure for CEMNet to make it work with complex inputs.
	
	(3) To the best of our knowledge, CEMNet is the first model trained in the frequency domain that can achieve more than 70\% validation accuracy on CIFAR-10 database.

	\section{Mathematical Basis}
	\label{Math}
	
	In the field of signal processing, {\bfseries Convolution Theorem} is one of the basic calculation rule. Assuming that ${\bf u}$ and ${\bf v}$ are two casual signals in the time domain, $\ast$ the convolution operation, and $\mathcal{F}(\cdot)$ the Discrete Fourier transform (DFT), then we have:
	
	\begin{equation}
		\label{eq:ct}
		\mathcal{F}({\bf u} \ast {\bf v}) = \mathcal{F}({\bf u}) \cdot \mathcal{F}({\bf v})
	\end{equation}
	
	where $\cdot$ is the element-wise multiplication. 
	
	Several previous studies consider to apply the Convolution Theorem on CNNs to transfer the complicate image convolution operation to the simple element-wise multiplication. However, the so-called image convolution in CNNs is in fact the cross-correlation operation in signal processing. Thus we need to consider the {\bfseries Cross-Correlation Theorem}:
	
	\begin{equation}
		\label{eq:cct}
		\mathcal{F}(R({\bf u}, {\bf v})) = \mathcal{F}^*({\bf u}) \cdot \mathcal{F}({\bf v})
	\end{equation}
	
	where $R({\bf u}, {\bf v})$ is the cross-correlation between ${\bf u}$ and ${\bf v}$, and $\mathcal{F}^*({\bf u})$ is the conjugate complex number of $\mathcal{F}({\bf u})$. 
	
	Eq.~\ref{eq:cct} serves as an important basis of our research since it builds a relationship between the time domain and frequency domain for convolution layers in CNNs. 
	
	\section{Method}
	\label{method}
	
	In this section, we firstly introduce necessary network layers in the frequency domain, then propose the implementation method of CEMNet.
	
	\subsection{Important Layers}
	
	\subsubsection{Element-wise Multiplication Layer}
	
	Based on Section~\ref{Math}, the image convolution operation in the time domain can be replaced by using element-wise multiplication in the frequency domain. Therefore, we design the element-wise multiplication layer as the most important part of CEMNet. 
	
	Assuming that we have a convolution layer $L$, which has $H_1 \times H_2 \times C_{in} $ sized input feature map ${\bf I}_L$ and $H_1 \times H_2 \times C_{out} $ sized output feature map ${\bf O}_L$. Notice that here we only consider the situation that ${\bf I}_L$ and ${\bf O}_L$ have the same $H_1$ and $H_2$. The convolution filter ${\bf W}$ of $L$ has the size of $K \times K \times C_{in} \times C_{out}$, where $K \le H_1$ and $K \le H_2$. Thus the forward process of $L$ is :
	
	\begin{equation}
		\label{convforward}
		{\bf O}_L = {\bf I}_L \ast {\bf W}
	\end{equation}
	
	where $\ast$ is the image convolution operation (which is in fact the cross-correlation in signal processing). It is easy to know that the computation complexity of one convolution layer is $O(K^2 \times H_1 \times H_2 \times C_{in} \times C_{out})$. According to Eq.~\ref{eq:cct}, the corresponding operation of Eq.~\ref{convforward} in the frequency domain is:
		
	\begin{equation}
		\label{eq:convforwardfreq}
		\mathcal{F}({\bf O}_L) = \mathcal{F}^*({\bf I}_L) \cdot \mathcal{F}({\bf W}_p)
	\end{equation}
	
	Notice that we should firstly do zero-padding on ${\bf W}$ to generate the padded filter ${\bf W}_p$ to guarantee that it has the same height and width as ${\bf I}_L$. In our research, we simply put ${\bf W}$ at the upper-left corner of ${\bf W}_p$ and set the rest elements of ${\bf W}_p$ to $0$. Moreover, we should expand $\mathcal{F}^*({\bf I}_L)$ to $H_1 \times H_2 \times C_{in} \times C_{out}$ by simply copy it for $C_{out}$ times. We can easily find that the computation complexity of one Element-wise Multiplication Layer is reduced to $O(H_1 \times H_2 \times C_{in} \times C_{out})$. At the end of the calculation, we sum over the third dimension of $\mathcal{F}({\bf O}_L)$ to make it has the size of $H_1 \times H_2 \times C_{out} $.
	
	Eq.~\ref{eq:convforwardfreq} is the forward calculation of the proposed element-wise multiplication layer, and it is very easy to derive the gradient of the layer:
	
	\begin{equation}
		\label{eq:convbackwardfreq}
		\frac{\partial \mathcal{F}({\bf O}_L)}{\partial \mathcal{F}({\bf W}_p)} = \mathcal{F}^*({\bf I}_L)
	\end{equation}
	
	Eq.~\ref{eq:convbackwardfreq} implies that the gradient calculation of element-wise multiplication layers is much easier than regular convolution layers.
	
	Unfortunately, if we use the above-mentioned equations directly to implement an element-wise multiplication layer, it may not result in a good performance. The main reason for this fact can be attributed to the over-fitting problem. CNNs use local receptive fields to reduce the number of free parameters, thus in most cases, convolution filters tend to be much smaller than input features. For instance, a convolution layer with $32 \times 32 \times 3$ sized ${\bf I}_L$ may only need to use a $3 \times 3 \times 3 \times 64$ sized ${\bf W}$ to output $32 \times 32 \times 64$ sized ${\bf O}_L$, and in this case the number of free parameters is only $1728$. However, if we transfer this convolution layer into the frequency domain, the scale of $\mathcal{F}({\bf W}_p)$ becomes $32 \times 32 \times 3 \times 64$, which corresponds to $196,608$ free parameters. The sharply increased free parameters mean that the over-fitting may happen much easier on CEMNets. To fix this problem, we introduce a {\bfseries Weight Fixation} mechanism during the training process. Specifically, $\mathcal{F}({\bf W}_p)$ should be transferred back to the time domain (still denoted by ${\bf W}_p$) after each weight update. After that, we perform an element-wise multiplication among ${\bf W}_p$ and a Weight Fixation matrix ${\bf V}$, where ${\bf V}$ is a $0-1$ matrix and only upper-left $K \times K$ elements are set to $1$. Thus Eq.~\ref{eq:convforwardfreq} becomes:
	
	\begin{equation}
		\label{eq:convforwardfreqwithfix}
		\mathcal{F}({\bf O}_L) = \mathcal{F}^*({\bf I}_L) \cdot \mathcal{F}({\bf W}_p \cdot {\bf V})
	\end{equation}
	
	The computation complexity of DFT for Weight Fixtation is $O(H_1 \times H_2 \times C_{in} \times C_{out} \times log(H_1 \times H_2))$, and we can learn that in most cases the computation complexity of an element-wise multiplication layer with Weight Fixation is still lower than the corresponding convolution layer. Even though the proposed Weight Fixation mechanism slows down the training process, it can obviously improve network performance.

	One important drawback of the above-mentioned element-wise multiplication layer is the high memory usage, especially for the large-scale databases like ImageNet. For instance, considering the first convolution layer of a network for ImageNet classification: the input images have the size of $224$-by-$224$, while the convolution filter has the size of $3$-by-$3$-by-$3$-by-$64$. In this case, the corresponding weight matrix in the frequency domain is $224$-by-$224$-by-$3$-by-$64$, which costs about $5575$ times memory of the time domain counterpart. To reduce the memory usage of CEMNets, we only store the time domain convolution filters of each element-wise multiplication layers. At the beginning of each training step, we transfer the convolution filters to the frequency domain and perform forward and backward calculations. Then we use the gradients to update $\mathcal{F}({\bf W}_p)$, then transfer them back to the time domain. Notice that the weight fixation is performed at the same time. In this way, the memory cost of the CEMNet can be obviously reduced. 
	
	\subsubsection{Batch Normalization}
	
	Batch Normalization \cite{ioffe15} is a widely-used regularization method in deep learning. By normalizing each training mini-batch, Batch Normalization, to a large extent, solves the problem of internal covariance shift, and improves the stability of the training process. Assuming that ${\bf B} = \{{\bf u}_1, {\bf u}_2, ..., {\bf u}_S\}$ is a training mini-batch with batch size $S$, the basic procedure of Batch Normalization can be divided into two steps:
	
	1. Normalization: we should firstly calculate the mean $\mu_{\bf B}$ and variance $\sigma_{\bf B}$ over the mini-batch ${\bf B}$, then normalize training samples in ${\bf B}$:
	\begin{equation}
		\label{eq:normalization}
		\hat{{\bf u}}_i = \frac{{\bf u}_i - \mu_{\bf B}}{\sqrt{\sigma_{\bf B}^2 + \epsilon}}, i = 1, ..., S
	\end{equation}
	
	where $\epsilon$ is a small enough constant to prevent zero-denominator. 
	
	2. Scale and shift:  in this step two learnable parameters $\gamma$ and $\beta$ are introduced to perform the Batch Normalization Transform on $\hat{x_i}$:
	\begin{equation}
		\label{eq:bnormtransform}
		BNT_{\gamma, \beta}({\bf u}_i) = \gamma \hat{{\bf u}_i} + \beta, i = 1, ..., S
	\end{equation}
	
	Based on the definition of Discrete Fourier Transform, we transfer time domain features to the frequency domain follow the Eq.~\ref{eq:DFT}:
	
	\begin{equation}
		\label{eq:DFT}
		\begin{split}
			& \mathcal{F}_{real}(u,v) = \sum_{x=0}^{M-1}\sum_{y=0}^{N-1} f(x,y) \cos(2\pi(\frac{ux}{M}+\frac{vy}{N}))  \\
			& \mathcal{F}_{imag}(u,v) = -\sum_{x=0}^{M-1}\sum_{y=0}^{N-1} f(x,y) \sin(2\pi(\frac{ux}{M}+\frac{vy}{N})) \\
			& u = 0, 1, ..., M-1;   v = 0, 1, ..., N-1
		\end{split}
	\end{equation}
	
	where $\mathcal{F}_{real}(u,v)$ and $\mathcal{F}_{imag}(u,v)$ are the real part and imaginary part of the features in the frequency domain respectively. If we perform Batch Normalization on the time domain, then $\mathcal{F}_{real}(u,v)$ becomes:
	
	\begin{equation}
		\label{eq:DFTreal}
		\begin{split}
			& \mathcal{F}_{i,real}^{BNT}(u,v) = \sum_{x=0}^{M-1}\sum_{y=0}^{N-1} (\gamma \frac{f_i(x,y)-\mu_{\bf B}}{\sqrt{\sigma_{\bf B}^2 + \epsilon}} + \beta) \cos(2\pi(\frac{ux}{M}+\frac{vy}{N}))  \\
			& = \frac{\gamma}{\sqrt{\sigma_{\bf B}^2 + \epsilon}} (\sum_{x=0}^{M-1}\sum_{y=0}^{N-1} (f_i(x,y)-\mu_{\bf B}) \cos(2\pi(\frac{ux}{M}+\frac{vy}{N}))) + \sum_{x=0}^{M-1}\sum_{y=0}^{N-1} \beta \cos(2\pi(\frac{ux}{M}+\frac{vy}{N})) \\
			& = \gamma \frac{\mathcal{F}_{i,real}(u,v) - \sum_{x=0}^{M-1}\sum_{y=0}^{N-1} \mu_{\bf B} \cos(2\pi(\frac{ux}{M}+\frac{vy}{N}))}{\sqrt{\sigma_{\bf B}^2 + \epsilon}} + \beta_{real}, i = 1, ..., S
		\end{split}
	\end{equation}
	
	where $\gamma$ and $\beta_{real}$ can be viewed as a learnable parameter. Moreover, we have:
	
	\begin{equation}
		\label{eq:mureal}
		\begin{split}
			& \mathcal{F}_{real}(\mu_{\bf B})  = \sum_{x=0}^{M-1}\sum_{y=0}^{N-1} (\frac{1}{K} \sum_{i=1}^{k} f_i(x,y))  \cos(2\pi(\frac{ux}{M}+\frac{vy}{N})) \\
			& = \frac{1}{K} \sum_{i=1}^{k} \sum_{x=0}^{M-1}\sum_{y=0}^{N-1} f_i(x,y) \cos(2\pi(\frac{ux}{M}+\frac{vy}{N})) = \frac{1}{K} \sum_{i=1}^{k} \mathcal{F}_{i,real}(u, v)  = \mu_{\mathcal{F}, real} \\	
		\end{split}
	\end{equation}
	
	therefore, Eq.~\ref{eq:DFTreal} becomes:
	
	\begin{equation}
		\label{eq:DFTreal2}
		\begin{split}
			& \mathcal{F}_{i,real}^{BNT}(u,v) = \gamma \frac{\mathcal{F}_{i,real}(u,v) - \mu_{\mathcal{F}, real}}{\sqrt{\sigma_{\bf B}^2 + \epsilon}} + \beta_{real}, i = 1, ..., S
		\end{split}
	\end{equation}
	
	Similarly, we can get the corresponding $\mathcal{F}_{i,imag}^{BNT}(x,y)$ as below:
	
	\begin{equation}
		\label{eq:DFTimag}
		\begin{split}
			& \mathcal{F}_{i,imag}^{BNT}(u,v) = \gamma \frac{\mathcal{F}_{i,imag}(u,v) - \mu_{\mathcal{F}, imag}}{\sqrt{\sigma_{\bf B}^2 + \epsilon}} + \beta_{imag}, i = 1, ..., S
		\end{split}
	\end{equation}
	
	It is easy to find that Eq.~\ref{eq:DFTreal2} and Eq.~\ref{eq:DFTimag} have similar form as the function of Batch Normalization in Eq.~\ref{eq:bnormtransform}. The main difference is that $\mathcal{F}_{i,real}^{BNT}(u,v)$ and $\mathcal{F}_{i,imag}^{BNT}(u,v)$ share the variance of the time domain $\sigma_{\bf B}^2$ as the denominator. Considering the variance of real part and imaginary part over each training mini-batch in the frequency domain, which can be denoted as $\sigma_{F, real}^2$ and $\sigma_{F, imag}^2$. Since $\sigma_{\bf B}^2$, $\sigma_{\mathcal{F}, real}^2$ and $\sigma_{\mathcal{F}, imag}^2$ are constant for each mini-batch, we can assume that $\frac{\sigma_{\bf B}^2}{\sigma_{\mathcal{F}, real}^2} = C_{real}$ and $\frac{\sigma_{\bf B}^2}{\sigma_{\mathcal{F}, imag}^2} = C_{imag}$, where $C_{real}$ and $C_{imag}$ are constants. Because $\epsilon$ is a negligible constant, Eq.~\ref{eq:DFTreal2} and Eq.~\ref{eq:DFTimag} can be re-written as:
	
	\begin{equation}
		\label{eq:DFTreal3}
		\begin{split}
			& \mathcal{F}_{i,real}^{BNT}(u,v) = \frac{\gamma}{\sqrt{C_{real}}} \frac{\mathcal{F}_{i,real}(u,v) - \mu_{\mathcal{F}, real}}{\sqrt{\sigma_{\mathcal{F}, real}^2 + \epsilon}} + \beta_{real}, i = 1, ..., S
		\end{split}
	\end{equation}
	
	and 
	
	\begin{equation}
		\label{eq:DFTimag2}
		\begin{split}
			& \mathcal{F}_{i,imag}^{BNT}(u,v) = \frac{\gamma}{\sqrt{C_{imag}}} \frac{\mathcal{F}_{i,imag}(u,v) - \mu_{\mathcal{F}, imag}}{\sqrt{\sigma_{\mathcal{F}, imag}^2 + \epsilon}} + \beta_{imag}, i = 1, ..., S
		\end{split}
	\end{equation}
	
	where $\frac{\gamma}{\sqrt{C_{real}}}$ and $\frac{\gamma}{\sqrt{C_{imag}}}$ can be viewed as learnable parameters.
	
	Based on the analyses above, we can learn that the implementation of Batch Normalization in the frequency domain has exactly the same form as the time domain. Thus in practice, we can directly do Batch Normalization on the real part and imaginary part of complex features in each mini-batch respectively. 
	
	\subsubsection{Approximated Dropout}
	
	Dropout \cite{srivastava2014dropout} is another widely-used regularization method for deep neural networks. By randomly dropping part of neurons during the training process, Dropout obviously relieves the over-fitting problem and improves the generalization ability of neural networks. Specifically, every neuron of the network may be dropped with probability $p$ during the training process. Assuming that $f(x,y)$ is one of the neurons of the time domain feature. Then the Dropout can be written as $f_d(x, y) = r f(x, y)$, where $r$ follows the Bernoulli distribution with probability $1-p$ (which means the Dropout rate is $p$).
	

%
	Based on Eq.~\ref{eq:DFT} and the definition of Dropout, we can learn that if we perform Dropout on the time domain, then the corresponding frequency domain feature becomes:
	
	\begin{equation}
		\label{eq:DFTdropped}
		\begin{split}
			& \mathcal{F}_{d,real}(u,v) = \sum_{x=0}^{M-1}\sum_{y=0}^{N-1} r(x,y) f(x,y) \cos(2\pi(\frac{ux}{M}+\frac{vy}{N}))  \\
			& \mathcal{F}_{d,imag}(u,v) = -\sum_{x=0}^{M-1}\sum_{y=0}^{N-1} r(x,y) f(x,y) \sin(2\pi(\frac{ux}{M}+\frac{vy}{N}))
		\end{split}
	\end{equation}
	
	from Eq.~\ref{eq:DFTdropped} we can learn that it is almost impossible that $\mathcal{F}_{d,real}(u,v)$ and $\mathcal{F}_{d,imag}(u,v)$ equal to $0$. Thus we can not directly use the regular Dropout on the frequency domain features like Batch Normalization. Moreover, based on Eq.~\ref{eq:convforwardfreq} we can know that if we directly set part of neurons in $F({\bf I}_L)$ to $0$, the corresponding neurons in $F({\bf O}_L)$ also become $0$. With the network going deep, an increasing proportion of neurons may die, which prevents the network from getting good performance. 
	
	To implement Dropout in the frequency domain, we propose an approximation method. The foundation of our approximation method is the observation that performing Dropout on $f(x, y)$ equals to randomly shrink or amplify $\mathcal{F}_{d,real}(u,v)$ and $\mathcal{F}_{d,imag}(u,v)$, because the values of both $\cos(2\pi(\frac{ux}{M}+\frac{vy}{N}))$ and $\sin(2\pi(\frac{ux}{M}+\frac{vy}{N}))$ have the same probability that within the ranges of $[-1, 0]$ and $(0, 1]$. Therefore, we have:
	
	\begin{equation}
		\label{eq:DFTdroppedappr}
		\begin{split}
			& \mathcal{F}_{d,real}(u,v) \simeq r_{appr,real}(u,v) \mathcal{F}_{real}(u,v) \\
			& \mathcal{F}_{d,imag}(u,v) \simeq r_{appr,imag}(u,v) \mathcal{F}_{imag}(u,v)
		\end{split}
	\end{equation}
	
	Based on the analyses above, we make an assumption that $r_{appr,real}(u,v)$ and $r_{appr,imag}(u,v)$ should not have large deviation from $1$, since all elements of the time domain features have the same probability to be dropped. Assuming that the mean of $f(x, y)$ is $\mu_f$, then the expectation of the mean of $r(x, y)f(x, y)$ is $(1-p)\mu_f$. For $\mathcal{F}_{d,real}(u,v)$, since the upper bound of $\cos(2\pi(\frac{ux}{M}+\frac{vy}{N}))$ is $1$ while the lower bound is $-1$, it is easy to learn that $r_{appr,real}(u,v)$ may have higher probability to lay between $1-p$ and $1+p$. Similarly, $r_{appr,imag}(u,v)$ may also lay between $1-p$ and $1+p$. 
	
	Therefore, we make $r_{appr,real}(u,v)$ and $r_{appr,imag}(u,v)$ obey a normal distribution $\mathcal{N}(1, p/2)$. Here we set the mean of the Normal distribution equals to $1$ and standard deviation $p/2$. Thus the probability that $r_{appr,real}(u,v)$ and $r_{appr,imag}(u,v)$ lay between $1-p$ and $1+p$ is about $95.4\%$. Notice that we ignore the relationship between $\cos(2\pi(\frac{ux}{M}+\frac{vy}{N}))$ and $\sin(2\pi(\frac{ux}{M}+\frac{vy}{N}))$ to simplify the calculation, thus $r_{appr,real}(u,v)$ and $r_{appr,imag}(u,v)$ are independent with each other. In practice we set $p=0.5$ for all layers, and the proposed approximated Dropout shows good performance in the frequency domain.

	

	\subsubsection{Approximated Leaky ReLU}
	
	Currently, rectified linear unit (ReLU) \cite{dahl2013improving} is a prevalent activation function. The form of ReLU is $f(x) = \max (0, x)$, which is very simple, and may increase the training speed of neural networks because its gradient can be calculated much easier. Leaky ReLU \cite{Maas2013RectifierNI} is a generalization of ReLU, which can be defined as below:
	
	\begin{equation}
		\label{LReLU}
		f(x) =
		\begin{cases}
			x&   x \ge 0, \\
			ax&   x < 0.
		\end{cases}
	\end{equation}
	
	where $a$ is a pre-defined constant. In this way, Leaky ReLU may avoid the 'dead ReLU' problem, which means a large number of neurons cannot be activated since they have always been set to $0$ by using ReLU. If we perform Leaky ReLU on the time domain feature, then the corresponding frequency domain feature is:
	
	\begin{equation}
		\label{eq:DFTLReLU}
		\begin{split}
			& \mathcal{F}_{lr,real}(u,v) = \sum_{x=0}^{M-1}\sum_{y=0}^{N-1} LReLU(f(x,y)) \cos(2\pi(\frac{ux}{M}+\frac{vy}{N}))  \\
			& \mathcal{F}_{lr,imag}(u,v) = -\sum_{x=0}^{M-1}\sum_{y=0}^{N-1} LReLU(f(x,y)) \sin(2\pi(\frac{ux}{M}+\frac{vy}{N}))
		\end{split}
	\end{equation}
	
	Similar to Dropout, we can easy to learn that doing Leaky ReLU on the time domain equals to shrink or amplify the value of both $\mathcal{F}_{real}(u,v)$ and $\mathcal{F}_{imag}(u,v)$. Even though this operation is not a random process, we may still apply a normal distribution to simulate it in the frequency domain since the locations of negative elements in the time domain features can be viewed as random:
	
	\begin{equation}
		\label{eq:DFTLReLUappr}
		\begin{split}
			& \mathcal{F}_{lr,real}(u,v) \simeq \mathcal{N}(\mu_{real}, \sigma_{real}) \mathcal{F}_{real}(u,v) \\
			& \mathcal{F}_{lr,imag}(u,v) \simeq \mathcal{N}(\mu_{imag}, \sigma_{imag}) \mathcal{F}_{imag}(u,v)
		\end{split}
	\end{equation}

    Follow the analyses of approximated Dropout, we can learn that  $\mu_{real}$ and $\mu_{imag}$ can be set to $1$, while $\sigma_{real}$ and $\sigma_{imag}$ are unknown but should not have a large value. In practice we use $p/4$ as the $\sigma_{real}$ and $\sigma_{imag}$ to guarantee that the randomness of approximated Leaky ReLU layers less then approximated Dropout layers.

    One potential problem of the proposed approximated Leaky ReLU is the random errors introduced by normal distributions in the frequency domain. Fortunately, we always make Leaky ReLU layers and Dropout layers appear together in the time domain. Therefore, the corresponding operation in the frequency domain can be viewed as a multiplication of two normal distributions:
    
    \begin{equation}
    	\label{eq:droplreluappr}
    	\begin{split}
    		& \mathcal{D}_{real} = \mathcal{N}(1, \sigma_{real}) \times \mathcal{N}(1, p/2) \\
    		& \mathcal{D}_{imag} = \mathcal{N}(1, \sigma_{imag}) \times \mathcal{N}(1, p/2)
    	\end{split}
    \end{equation}

    then we have:
    
    \begin{equation}
    	\label{eq:droplreluappr2}
    	\begin{split}
    		& \mathcal{D}_{real} \simeq \frac{1}{\sqrt{2\pi (\sigma_{real}^2 + (p/2)^2)}} \mathcal{N}(1, \sqrt{\frac{\sigma_{real}^2 (p/2)^2}{\sigma_{real}^2 + (p/2)^2}}) \\
    		& \mathcal{D}_{imag} \simeq \frac{1}{\sqrt{2\pi (\sigma_{imag}^2 + (p/2)^2)}} \mathcal{N}(1, \sqrt{\frac{\sigma_{imag}^2 (p/2)^2}{\sigma_{imag}^2 + (p/2)^2}})
    	\end{split}
    \end{equation}

    Eq.~\ref{eq:droplreluappr2} shows that $\mathcal{D}_{real}$ and $\mathcal{D}_{imag}$ follow scaled normal distributions respectively. Since we set $\sigma_{real}$ and $\sigma_{imag}$ to $p/4$, it is highly possible that the approximated Dropout layers can cover the random errors brought by the approximated Leaky ReLUs. Thus Eq.~\ref{eq:droplreluappr2} becomes:
    
    \begin{equation}
    	\label{eq:droplreluappr3}
    	\begin{split}
    		& \mathcal{D}_{real} \simeq \frac{2}{\sqrt{5} \pi p} \mathcal{N}(1, \frac{p}{2\sqrt{5}}) \\
    		& \mathcal{D}_{imag} \simeq \frac{2}{\sqrt{5} \pi p} \mathcal{N}(1, \frac{p}{2\sqrt{5}})
    	\end{split}
    \end{equation}
	
	Even though the working behavior of the proposed approximation method is not exactly the same as Leaky ReLU in the time domain, the subsequent normal distribution, which is applied to approximate Dropout, can cover the deviations with high probability. Thus the proposed approximated Leaky ReLU and approximated Dropout in the frequency domain can be viewed as a simulation of the combination of a Leaky ReLU layer and a Dropout layer with a variable dropout rate in the time domain. 
	
	
	\subsubsection{Max Pooling}
	
	Max Pooling is an essential element of CNNs, since it serves as a tool for down-sampling and has good performance in many databases and real-world tasks. Unfortunately, Max Pooling cannot work in the frequency domain since complex numbers cannot compare with each other. Therefore, to implement pooling layers of CEMNet, we firstly transfer the input feature maps back to the time domain, then perform Max Pooling and transfer the results back to the frequency domain. The computation complexities of the DFT and iDFT operation here are $O(H_1 \times H_2 \times C_{in} \times log(H_1 \times H_2))$ and $O(H_1^{'} \times H_2^{'} \times C_{out} \times log(H_1^{'} \times H_2^{'}))$, respectively. Where $H_1^{'}$ and $H_2^{'}$ are the size of down-sampled features.  
	
	\subsection{Complex Element-wise Multiplication Network}
	
	Based on the analysis of important layers above, we found that we can process real parts and imaginary parts of input features separately for most kinds of layers. Inspired by \cite{Guberman16ComplexCNN} and \cite{trabelsi2018deep}, we design a two-branches structure of CEMNet to integrate all network layers in the frequency domain. 
	
	The basic idea of CEMNet is to introduce two branches (one for the real part and the other for the imaginary part of complex features) in each layer. For instance, the implementation of Element-wise Multiplication Layers (see Eq.~\ref{eq:convforwardfreqwithfix}) can be described as Algorithm~\ref{alg_EML}. We can use a similar method to implement the other layers. It is easy to know that in Element-wise Multiplication Layers we use $4$ real-value multiplication to replace $1$ complex-value multiplication, while in the rest two-branches layers we use $2$ real-value multiplication to replace $1$ complex-value multiplication.
	
	\begin{algorithm}[h]
		\caption{The operations of the two-branches Element-wise Multiplication Layer}
		\begin{algorithmic}  \label{alg_EML}
			\STATE {\bfseries Input:} The real part of input feature map $\mathcal{F}_{real}^*({\bf I})$, the imaginary part of input feature map $\mathcal{F}_{imag}^*({\bf I})$, the real part of weight $\mathcal{F}_{real}({\bf W})$, and the imaginary part of weight $\mathcal{F}_{imag}({\bf W})$
			
			\STATE {\bfseries Output:} The real part of output feature map $\mathcal{F}_{{\bf O}, real}$, and the imaginary part of output feature map $\mathcal{F}_{{\bf O}, imag}$
			
			\STATE Do Weight Fixation on ${\bf W}$ to update $\mathcal{F}_{real}({\bf W})$ and $\mathcal{F}_{imag}({\bf W})$
			
			\STATE ${\bf F}_{rr} = \mathcal{F}_{real}^*({\bf I}) \cdot \mathcal{F}_{real}({\bf W})$, ${\bf F}_{ri} = \mathcal{F}_{real}^*({\bf I}) \cdot \mathcal{F}_{imag}({\bf W})$, ${\bf F}_{ir} = \mathcal{F}_{imag}^*({\bf I}) \cdot \mathcal{F}_{real}({\bf W})$, and ${\bf F}_{ii} = \mathcal{F}_{imag}^*({\bf I}) \cdot \mathcal{F}_{imag}({\bf W})$. Where $\cdot$ is element-wise production
			
%
%
			
			\STATE $\mathcal{F}_{{\bf O}, real} = LeakyReLU(BNorm({\bf F}_{rr} + {\bf F}_{ii}))$
			
			\STATE $\mathcal{F}_{{\bf O}, imag} = LeakyReLU(BNorm({\bf F}_{ri} - {\bf F}_{ir}))$
			
		\end{algorithmic}
	\end{algorithm}

	At the end of CEMNet, we flatten the real part and imaginary part of complex feature maps, then feed them into one or more regular fully connected layers, respectively. Finally, we concatenate the real feature vectors and imaginary feature vectors, and use one fully connected layer to generate classification results. 
	
	
	\section{Experimental Results} 
	\label{experiments}
	
	\subsection{Databases and Computation Platform}
	
	In this paper, we apply two widely used databases to evaluate the proposed CEMNet, i.e., MNIST and CIFAR-10. MNIST database \cite{lecun1998gradient} is a hand-written digits database that contains 60,000 training images and 10,000 test images, and the images are 28-by-28 in size. All images are grey-scale and only contain one digit. CIFAR-10 database \cite{krizhevsky2014cifar} is widely used in image classification tasks. It contains 50,000 32-by-32 RGB images for training and 10,000 images for validation, and all images are divided into 10 classes. We implement the proposed CEMNet on Tensorflow 2.5 \cite{tensorflow2015-whitepaper} platform, and our computation platform includes Intel Xeon 4108 CPU, 128 GB memory, and two Tesla V100 GPUs (for large CEMNet we use 6 Tesla A40 GPUs){\footnote{The url of our source codes: https://github.com/mowangphy88/CEMNet}}.
	
	\subsection{MNIST Experiments}
	
	The baseline CNN structure of MNIST experiments is based on LeNet-5 \cite{LeNet}. Comparing with the original LeNet-5, the feature map size of the second convolution layer of our network is set to $14 \times 14$ instead of $10 \times 10$ by using zero-padding, thus after the second down-sampling, the feature map size becomes $7 \times 7$. The activation function of each layer is Leaky-ReLU with $a=0.2$. Moreover, we add Batch Normalization and Dropout after each convolution layer and fully-connected layer. Based on the CNN structure, we build the corresponding CEMNet to test the performance in the frequency domain. 
	
	During the training process of CEMNet, we use RMSProp as the optimizer. The batch-size is set to $100$, and the network should be trained for $800$ epochs. We use a cosine decay learning rate schedule with an initial learning rate $= 0.004$ and a minimum learning rate $= 0.0000001$. To initialize the element-wise multiplication layers, we firstly use he\_normal \cite{he2015delving} to initialize corresponding convolution kernels, then transfer them to the frequency domain. We perform the Weight Fixation on all element-wise multiplication layers. Notice that we do not apply any data augmentation methods during the training process.
	
	We select several state-of-the-art frequency domain neural network methods as baselines to compare with CEMNet. Moreover, to show the importance of the proposed methods, we also include ablation studies. The experimental results of MNIST are included in Table~\ref{table-MNIST}, and we can learn that Weight Fixation, Batch Normalization, and Dropout work well in the frequency domain. Generally speaking, CEMNet has a better performance compared with previous DFT based methods, and comparable with its CNN counterpart. Moreover, the CEMNet has much less number of operations than the corresponding CNN. Notice that all DFTs and the extra operations introduced by the two-branches structure of the CEMNet are taken into account.
	
	\begin{table}[h]
		\vskip 0.1in
		\begin{center}
			\begin{small}
				\begin{sc}
					\caption{Comparison between different baselines and CEMNets on MNIST. (WF: Weight Fixation, BN: Batch Normalization. The paper of FCNN only provides learning curves instead of exact performance)}
					\label{table-MNIST}
					\begin{tabular}{c|c|c|c}
						\toprule
						Methods                             &   Forward Ops  &   Backward Ops & Test Error     \\
						\midrule
						Modified LeNet-5                    &  $\approx$ 692K   &    $\approx$ 692K    & 0.72\%        \\
						\midrule
						2SReLU \cite{WATANABE2021107851}     &    N/A        &      N/A          & 3.08\%        \\
						FCNN \cite{FCNN}                     &    N/A        &      N/A          & $\approx$ 3\%  \\
						\midrule
						CEMNet - WF - BN - Dropout           &    N/A        &     N/A               & 1.11\%        \\
						CEMNet - BN - Dropout               &    N/A         &    N/A                & 0.99\%         \\
						CEMNet - Dropout                     &    N/A        &    N/A                & 0.91\%         \\
						CEMNet                               &   $\approx$ 368K &    $\approx$ 481K   & 0.67\%         \\
						\bottomrule
					\end{tabular}
				\end{sc}
			\end{small}
		\end{center}
		\vskip -0.1in
	\end{table}
	
	\subsection{CIFAR-10 Experiments}
	
	The baseline CNN structure of CIFAR-10 experiments is based-on VGG-16 \cite{simonyan2014vgg} framework, which contains 5 blocks. Each block contains one or more convolution layers and ends with a sub-sampling layer, and the number of feature maps of convolution layers in each block is set to $64, 128, 256, 512, 512$. The network ends with one or more fully-connected layers. The activation function of each layer is Leaky-ReLU with $a=0.2$. We consider two scales of CNNs:
	
	(1) Small scale CNN: each block contains one convolution layer, and the network ends with one fully-connected layer with $512$ neurons.
	
	(2) Large scale CNN: the number of convolution layers in each block is set to $2, 2, 2, 3, 3$, and the network ends with one fully-connected layer with $512$ neurons.
	
	Notice that for both CNNs we also apply Batch Normalization and Dropout after each convolution layer and fully-connected layer. We build the corresponding CEMNets based on the above-mentioned two CNN structures. 
	
	The training process of CEMNets with CIFAR-10 database shares similar optimizer and initialization method with MNIST. The batch-size is set to $100$, and the network should be trained for $800$ epochs. We use a cosine decay learning rate schedule with an initial learning rate $= 0.004$ and a minimum learning rate $= 0.0000001$. We perform the Weight Fixation on the first $3$ blocks. Data augmentation methods are also not applied during the training process. Table~\ref{table-CIFAR} provides experimental results on CIFAR-10, which also show the advantages of the proposed method.
	
	
	From experiments of CIFAR-10 we can learn that the Small CEMNet has comparable performance to the corresponding Small CNN, but the performance gap between the Large CEMNet and Large CNN is relatively large. One potential reason for this situation is the Weight Fixation method. Even though the Weight Fixation reduces the number of free parameters and alleviate the problem of over-fitting, it discards a large amount of gradient information. Therefore, with the network going deep, the problem of gradient vanishing may happen much easier, which negatively impacts on the classification performance. 
	
	\begin{table}[h]
		\vskip 0.1in
		\begin{center}
			\begin{small}
				\begin{sc}
					\caption{Comparison between different baselines and CEMNets on CIFAR-10. (The paper of FCNN only provides learning curves instead of exact performance)}
					\label{table-CIFAR}
					\begin{tabular}{c|c|c|c}
						\toprule
						Methods                              & Forward Ops      & Backward Ops & Test Error     \\
						\midrule
						Small CNN                             &  $\approx$ 68.10M   &  $\approx$ 68.10M   & 21.07\%    \\
						Large CNN                             &  $\approx$ 275.45M  &  $\approx$ 275.45M  & 11.30\%    \\
						\midrule
						FCNN \cite{FCNN}                     &    N/A               &    N/A          & $\approx$ 73\%  \\
						\midrule
						Small CEMNet - WF - BN - Dropout       &    N/A              &    N/A          & 40.67\%        \\
						Small CEMNet - BN - Dropout             &    N/A             &    N/A            & 35.11\%         \\
						Small CEMNet - Dropout                  &    N/A              &    N/A            & 32.83\%         \\
						Small CEMNet                            &   $\approx$ 31.33M   &   $\approx$ 50.17M  & 22.40\%     \\
						Large CEMNet                            &   $\approx$ 124.39M  &   $\approx$ 194.28M  & 21.63\%       \\
						\bottomrule
					\end{tabular}
				\end{sc}
			\end{small}
		\end{center}
		\vskip -0.1in
	\end{table}

	\section{Conclusion and Future Works}
	
	In this paper, we propose a novel neural network model, namely CEMNet, that can be trained in the frequency domain. Based on the Cross-Correlation Theorem, we design the element-wise multiplication layer to remove all convolution operations from CNNs and reduce the computation complexity. To alleviate the problem of over-fitting, we introduce the Weight Fixation mechanism. Moreover, we theoretically analyze the working behavior of Batch Normalization, Leaky ReLU, and Dropout in the frequency domain, and introduce their counterparts into CEMNet. Also, to deal with complex inputs brought by DFT, we design a two-branches network structure for CEMNet. Experimental results show that CEMNets achieve better performance on MNIST and CIFAR-10 compared with previous DFT based methods. 
	
	This research still remains some open problems that should be solved in the future. The first problem is the network performance of large models. Based on our analyses in Section~\ref{experiments}, we can learn some potential problems of the Weight Fixation, and we may need to update this mechanism for better performance. Secondly, our method needs to repeatedly transfer data between the time domain and frequency domain, which may slow down the learning speed. We will consider to speed up our method by removing DFT operations during the training stage. 
	
	
	\bibliographystyle{plainnat} 
	\bibliography{ref}
	
	%
		%
		%
		%

\section*{Checklist}

The checklist follows the references.  Please
read the checklist guidelines carefully for information on how to answer these
questions.  For each question, change the default \answerTODO{} to \answerYes{},
\answerNo{}, or \answerNA{}.  You are strongly encouraged to include a {\bf
justification to your answer}, either by referencing the appropriate section of
your paper or providing a brief inline description.  For example:
\begin{itemize}
  \item Did you include the license to the code and datasets? \answerYes{See Section~\ref{gen_inst}.}
  \item Did you include the license to the code and datasets? \answerNo{The code and the data are proprietary.}
  \item Did you include the license to the code and datasets? \answerNA{}
\end{itemize}
Please do not modify the questions and only use the provided macros for your
answers.  Note that the Checklist section does not count towards the page
limit.  In your paper, please delete this instructions block and only keep the
Checklist section heading above along with the questions/answers below.

\begin{enumerate}

\item For all authors...
\begin{enumerate}
  \item Do the main claims made in the abstract and introduction accurately reflect the paper's contributions and scope?
    \answerYes{}
  \item Did you describe the limitations of your work?
    \answerYes{}
  \item Did you discuss any potential negative societal impacts of your work?
    \answerNA{}
  \item Have you read the ethics review guidelines and ensured that your paper conforms to them?
    \answerYes{}
\end{enumerate}

\item If you are including theoretical results...
\begin{enumerate}
  \item Did you state the full set of assumptions of all theoretical results?
    \answerNA{}
        \item Did you include complete proofs of all theoretical results?
    \answerNA{}
\end{enumerate}

\item If you ran experiments...
\begin{enumerate}
  \item Did you include the code, data, and instructions needed to reproduce the main experimental results (either in the supplemental material or as a URL)?
    \answerYes{}
  \item Did you specify all the training details (e.g., data splits, hyperparameters, how they were chosen)?
    \answerYes{}
        \item Did you report error bars (e.g., with respect to the random seed after running experiments multiple times)?
    \answerNA{}
        \item Did you include the total amount of compute and the type of resources used (e.g., type of GPUs, internal cluster, or cloud provider)?
    \answerYes{}
\end{enumerate}

\item If you are using existing assets (e.g., code, data, models) or curating/releasing new assets...
\begin{enumerate}
  \item If your work uses existing assets, did you cite the creators?
    \answerYes{}
  \item Did you mention the license of the assets?
    \answerNA{}
  \item Did you include any new assets either in the supplemental material or as a URL?
    \answerNA{}
  \item Did you discuss whether and how consent was obtained from people whose data you're using/curating?
    \answerNA{}
  \item Did you discuss whether the data you are using/curating contains personally identifiable information or offensive content?
    \answerNA{}
\end{enumerate}

\item If you used crowdsourcing or conducted research with human subjects...
\begin{enumerate}
  \item Did you include the full text of instructions given to participants and screenshots, if applicable?
    \answerNA{}
  \item Did you describe any potential participant risks, with links to Institutional Review Board (IRB) approvals, if applicable?
    \answerNA{}
  \item Did you include the estimated hourly wage paid to participants and the total amount spent on participant compensation?
    \answerNA{}
\end{enumerate}

\end{enumerate}


%
%
%
%

\end{document}